\documentclass{article}
\usepackage[utf8]{inputenc}
\usepackage{amsmath}
\usepackage{amsfonts}
\usepackage{graphicx}
\usepackage{hyperref}
\usepackage{authblk}

\title{Reconstruction of observed mechanical motions with Artificial Intelligence tools}
\author[1]{Antal Jakov\'ac\thanks{jakovac.antal@wigner.hu}}
\author[1]{Marcell T. Kurbucz\thanks{kurbucz.marcell@wigner.hu}}
\author[1]{P\'eter P\'osfay\thanks{posfay.peter@wigner.hu}}
\affil[1]{Wigner Research Centre for Physics, Department of Computational Sciences, 29-33 Konkoly-Thege Miklós Street, Budapest, H-1121, Hungary}
\date{\today}

\begin{document}
\maketitle

\setlength{\arraycolsep}{2pt}

\begin{abstract}
    The goal of this paper is to determine the laws of observed trajectories assuming that there is a mechanical system in the background and using these laws to continue the observed motion in a plausible way. The laws are represented by neural networks with a limited number of parameters. The training of the networks follows the Extreme Learning Machine idea. We determine laws for different levels of embedding, thus we can represent not only the equation of motion but also the symmetries of different kinds. In the recursive numerical evolution of the system, we require the fulfillment of all the observed laws, within the determined numerical precision. In this way, we can successfully reconstruct both integrable and chaotic motions, as we demonstrate in the example of the gravity pendulum and the double pendulum.
\end{abstract}

\section{Introduction}

The task of intelligent systems is to identify the phenomena or concepts that are the most appropriate to describe the observed data \cite{vervaeke, newell}. While these concepts are hard to approach in tasks like image recognition or text analysis, in the case of description of observed motions of dynamical systems, we have some clues of the general form of these concepts. These include that in a proper (phase) space the motion is governed by a first-order differential equation, or, in the case of continuum mechanics, first-order partial differential equations. Sometimes we also have some educated guess about the form of the kernel of the differential equation. But in complicated systems, the kernel must be determined from the observed data (data-driven modeling, c.f. \cite{datadriven}).

In these cases, the task of the (artificial) intelligence is to describe the kernel in the most reliable, and in the most sparse way. There are several approaches in the literature to do this. In the neural network-based approaches, one tries to set up a network that can learn the properties of the observed data. It is used for a wide range of fields, including general PDE \cite{Lietal,Umetal}, fluid dynamics \cite{Brenner,Bruntonetal,DaiSeljak}, quantum mechanics \cite{Snyderetal}, molecular dynamics \cite{Jacobsenetal}, particle dynamics \cite{PNN} and in chaotic systems \cite{chaotic, SINDy}.

In the course of reconstructing the underlying laws for an observed mechanical motion, we have to model the mathematics of the dynamical system. One can model the driving force of the system directly \cite{PNN, SINDy}, focus on the Hamiltonian \cite{Greydanusetal,Tothetal}, or the Lagrangian \cite{DaiSeljak, Cranmeretal}. The applied machine learning tool exhibits the functional space, where the best fit for the aimed function can be found. A very important problem in all of these approaches is to find the most robust approximation that respects the symmetries of the dynamics.

Here a new way is presented for the  reconstruction of the observed trajectories of simple and chaotic dynamical systems, using the Extreme Learning Machine \cite{Huangetal,Wangetal}. In this approach, only a part of the weights is trained in a neural network, usually the ones in  the last layer. This leads to a tremendous improvement both in required computational resources and training time.

The goal of this paper is to observe a motion with $\Delta t$ time separations, resulting in $x_n\in V$ series, where $V$ is some vector space. We want to determine a recursion kernel in order to represent the motion as an $x_{n+1} = F_{\Delta t}[x]$, where $[x]$ means the past; in mechanical systems, it is enough to consider the $x_n$ and $x_{n-1}$ values. We want to reconstruct the motion using this recursion and also continue it for future times.

An advantage of the above second-order recursion is that it, in principle, is appropriate to describe non-conservative systems, where the energy is not conserved. A drawback, however, 
is that a numerically determined force is improbable to support an exact conservation law even if it is conservative. This is because of observational noise and also reconstruction inaccuracies. In these systems, the recursion of the equation of motion soon leads to divergent trajectories, because usually the larger energy occupies larger phase space, and so by random walk, we tend to increase the system energy.

A solution to this problem follows the theoretical lines lied down in \cite{Jakovacetal}, and  applied to linear systems in \cite{Jakovaclinear}. We determine not only the equation of motion but also other "laws" of the systems, using different levels of embedding. A first-order embedding leads to a law of the form $C^{(1)}_{\Delta t}(x_n) = constant$, this includes the holonomic constraints of the system. Second-order constraints describe anholonomic constraints as well as other conserved quantities like energy. Third-order constraints give the equations of motion. In principle, higher-order constraints can be used too. The constraints are numerically determined and in the recursion the fulfillment of all of them is required within the precision of the numerical errors. This leads to a stable algorithm with good noise tolerance.

This paper has the following structure. In Section~\ref{sec:general} we discuss the general ideas to treat mechanical systems with finite time resolution. In Section~\ref{sec:numerics} we discuss the issues that come up in the course of a numerical representation. In Section~\ref{sec:studies} we turn to the actual studies: the mathematical pendulum, the physical pendulum, and the double pendulum. In Section~\ref{sec:conclusion} we close the paper with conclusions.

\section{General setup}
\label{sec:general}

In our earlier studies \cite{Jakovacetal, Jakovaclinear}, we established the multi-feature approach of AI and worked out the method for the determination of the linear laws. In this paper, it is examined how nonlinear laws can be treated.

The goal of this paper is that, assuming that we observe the motion of all degrees of freedom of a closed system, from the numerical data we determine the equations of motion, and continue the observed motion in a plausible way. In this sentence there are a lot of notions which have to be defined: how a motion is observed, what is understood under "all degrees of freedom" and under a "closed" system in general.

It is assumed that we observe a system $X$ that has possible states $x\in X$. We will represent the system as $X \equiv \mathbb R^N$. We record the state at every $t=n\Delta t$ time step, thus we have the $x_n$ records.

It is evident, that what happens tomorrow, must depend on what is present today, since forgotten things, which have no trace today, can not influence the fate of the system in the future. This means that we may set up a recursion
\begin{equation}
    \label{eq:recursion}
    x_{n+1} = F_{\Delta t}(x_n),\qquad \mathrm{where}\quad F_{\Delta t} : X \to X.
\end{equation}
The recursion also needs an initial condition $x_0$.

This form can be true only if all information that is necessary for the future of the system is accounted for. This means that "all degrees of freedom are accounted for".
Closedness is reflected in the fact that $F$ does not depend directly on time (i.e. on $n$).

If $\Delta t$ is small enough, then the numerical values of $x_{n+1}$ and $x_n$ are close to each other. Then it is usual to keep track only the change and define
\begin{equation}
    v_n = f_{\Delta t}(x_n),\qquad v_n = \frac{x_{n+1}-x_n}{\Delta t},\qquad f_{\Delta t}(x) =  \frac{F_{\Delta t}(x)-x}{\Delta t}.
\end{equation}

We shall emphasize here that $\Delta t$ is a finite quantity throughout the complete discussion. But we may discuss the $\Delta t\to0$ continuum limit. In this case
\begin{equation}
    v_n \stackrel{\Delta t\to0}{\longrightarrow} \dot x(t),\quad t = n\Delta t.
\end{equation}
Then the above recursion becomes a first-order differential equation
\begin{equation}
    \label{eq:diffeq_cont}
    \dot x = f_0(x),\quad x(t_0) = x_{init}.
\end{equation}
Its solution reads
\begin{equation}
    x(t) = G(t-t_0, x_{init}).
\end{equation}
$G$ depends only on $t-t_0$, if $f_0$ does not depend on time.

Choosing $t=(n+1)\Delta t$, $t_0= n\Delta t$ with a finite $\Delta t$, and $x_{init} = x_n$, we obtain
\begin{equation}
    x((n+1)\Delta t) = G(\Delta t, x(n\Delta t)).
\end{equation}
This is consistent with \eqref{eq:recursion} with $F_{\Delta t}(x) = G(\Delta t, x)$. This means that if a system is governed by an autonomous first-order differential equation, then the discrete-time evolution is governed by a first-order recursion.

Sometimes the equation \eqref{eq:diffeq_cont} has conserved quantities $C(x)$ which means that
\begin{equation}
    \label{eq:conserved_cont}
    C(x_{init}) = C(G(t-t_0, x_{init})),\qquad \forall x_{init}.
\end{equation}
This quantity is conserved for any $t-t_0$, thus this remains conserved in the discrete case, too.

In practice, it is not easy to decide, whether all information is known for the recursion, i.e. whether we know the complete state of the system. If the state space is $N$ dimensional ($X\sim \mathbb R^N$), then we know that we need $N$ real number to determine the future elements. Observing a single components of the embedded time series $x_0(t), \, x_0(t+\Delta t),\dots, \,x_0(t+(N-1)\Delta t)$ can provide this $N$ parameters: we have $N$ equations for $N$ unknowns for the components of $x_0$. Therefore it is also enough that we observe only some components of the complete state, but for several time steps, it then provides sufficient information to restore all the initial state, and so all the time dependence.

This means that we may observe a reduced system $y \in \mathbb R^M$, for time steps $t,\, t-\Delta t,\dots$. Then we consider the state of the system as
\begin{equation}
    x_n = \{ y_n, y_{n-1},\dots, y_{n-k} \}.
\end{equation}
This information is enough to provide $x_{n+1}$, and  $y_{n+1}$ can be computed. Thus we have the recursion
\begin{equation}
    y_{n+1} = \tilde F_{\Delta t}(y_n, y_{n-1},\dots, y_{n-k}).
\end{equation}
So in a reduced system, a $(k+1)$th order recursion is needed to describe the complete time evolution. In the continuum limit, it corresponds to a $(k+1)$th order differential equation. This statement is the difference equation version of Taken's embedding theorem \cite{takens}.

\subsection{Specialties of mechanical systems}

In mechanical systems, in the continuum limit, the states are elements of the phase space. If the configuration of the system is denoted by $x\in\mathbb R^N$, then the phase space consists of the $(x(t), \dot x(t))$ pairs.

In practice, we can only observe the configuration of the system. If all coordinates are observed, they are still needed at two different times. Therefore we should use the recursion
\begin{equation}
    \label{eq:Newton_discrete}
    x_{n+1} = F_{\Delta t}(x_n, x_{n-1}).
\end{equation}
It is worth introducing quantities to characterize the first- and second-order change
\begin{equation}
    \label{eq:va}
    v_n = \frac{x_{n}-x_{n-1}}{\Delta t},\qquad
    a_n = \frac{x_{n+1}-2x_n+x_{n-1}}{\Delta t^2},
\end{equation}
then we can write
\begin{equation}
    \label{eq:Newton}
    a_n = f_{\Delta t}(x_n,v_n),
\end{equation}
where $f_{\Delta t}(x,v)  =  [F_{\Delta t}(x, x-v\Delta t) -x-v\Delta t]/\Delta t^2$. This is the difference equation realization of the Newton-equation. $f_{\Delta t}$ is the discrete force function.

In the continuum limit, we have the differential equation
\begin{equation}
    \label{eq:Newton_cont}
    \ddot x = f_0(x, \dot x),
\end{equation}
an we have to fix the time evolution at two points, or at a single point we shall provide $x(t_0)$ and $\dot x(t_0)$. This fixes the solution uniquely:
\begin{equation}
    \label{eq:Newton_sol_cont}
    x(t) = G_{t_0-t_1}(t-t_0, x(t_0), x(t_1)).
\end{equation}
Because of time translation invariance this form depends only on the time differences. In particular with $t_0=n\Delta t$, $t_1=(n-1)\Delta t$ and $t=(n+1)\Delta t$ we get back \eqref{eq:Newton_discrete} with $F_{\Delta t}(x,x') = G_{\Delta t}(\Delta t, x, x')$.

We remark that even if $f_0$ does not depend on $\dot x$, i.e. the force is velocity independent, the discrete force $f_{\Delta t}$ can still depend on it.

We have seen in \eqref{eq:conserved_cont} that if in a continuum time there is a conserved quantity, then there is also in the discrete-time. Now we use the second derivative equation of motion, so the conserved quantities are $C(x,\dot x)$ in the continuum limit. Using \eqref{eq:Newton_sol_cont}, we can express the local velocity as a function of initial conditions:
\begin{equation}
    \dot x(t)=\dot G_{t_0-t_1}(t-t_0, x(t_0), x(t_1)).
\end{equation}
In particular with $t=t_0=n\Delta t$, $t_1=(n-1)\Delta t$
\begin{equation}
    \dot x(n\Delta t) = \dot G_{\Delta t}(0, x_n, x_{n-1}).
\end{equation}
This means that, choosing $t_0=(n-1)\Delta t$, $t_1=n\Delta t$
\begin{equation}
    C(x,\dot x)\bigr|_{t=n\Delta t} = C(x_n, \dot G_{\Delta t}(0, x_n, x_{n-1}) ) = \mathrm{constant}.
\end{equation}
This means that
\begin{equation}
    C_{\Delta t}(x,v) = C(x, \dot G_{\Delta t}(0, x, x-v\Delta t) )
\end{equation}
is a conserved quantity for all solutions of \eqref{eq:Newton} discrete recursion.

\section{The numerical problem}
\label{sec:numerics}

The problem we shall solve is that we observe some trajectories with fixed time resolution $\Delta t$, and we want to reproduce and also continue these trajectories. We emphasize that we want to reproduce the time series exclusively from observations, we shall not use the underlying differential equations. This also means that the local velocity information is not available, since it is not measured (only the discrete variant).

The algorithm to solve this problem is the following. We model the general function class to where it is suspected that our $f_{\Delta t}$ function belongs. This function class is approximated by a neural network with parameters $w \in \mathbb R^{N_w}$. Thus
\begin{equation}
    f_{\Delta t}(x,v) \approx {\cal F}(x,v ; w_{\Delta t}).
\end{equation}
The $w_i$ weights are determined to best describe the observed data. In very general terms we can say
\begin{equation}
    L_{total}(w) = \sum_{i,n} L(a^{(i)}_n,  {\cal F}(x^{(i)}_n,v^{(i)}_n ; w_{\Delta t})) = \mathrm{minimal},
\end{equation}
where $L(a,a')$ is the local loss, for example $L(a,a')=|a-a'|^2$; $L_{total}$ is the total loss, and $a^{(i)}_n$ and $v^{(i)}_n$ are the results of \eqref{eq:va} evaluated on $x^{(i)}_n$ trajectory.

In this paper, we use simulated data to train the network.

Once we have the approximate form of the discrete force function, we can establish the recursion
\begin{equation}
    \label{eq:recursion1}
    x_{n+1} = 2x_n + x_{n-1} + \Delta t^2 {\cal F}(x_n, \frac{x_x-x_{n-1}}{\Delta t} ; w_{\Delta t}).
\end{equation}

\subsection{The neural network}
\label{sec:network}

\begin{figure}[htbp]
    \begin{center}
        \includegraphics[height=4cm]{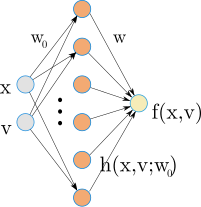}        
    \end{center}
    \caption{The neural network representation of the force function. $h(x,v;w_0)$ means the hidden layer values.}
    \label{fig:network}
\end{figure}

To approximate the discrete force we applied the Extreme Learning Machine \cite{Huangetal,Wangetal} ideas. This corresponds to a one hidden layer neural network, shown in Figure \ref{fig:network}. 
According to the logic of the Extreme Learning Machine, the $w_0$ weights in the first layer are not trained, they remain to be chosen randomly. Training concerns only the $w$ weights.

We can also say that from the input data we create random features at the hidden layer. From these random features, we compose the resulting function.

\subsubsection{Smoothness}

There are several issues we should take into account when we construct the neural network. The first one is smoothness. We expect that the discrete force is a smooth function of its arguments, meaning not just that there are no discontinuities present, but also that the derivative of the function should not exceed certain values. Physically we expect that nearby positions and velocities result in almost the same force.

Thus, we shall construct a network in which any reasonable choice of the weights results in an output which is a smooth function of the inputs. This requires that the features themselves are smooth for any $w_0$ values. In this work, we used the form
\begin{equation}
    h(x,v; w_0) = \frac1{ (x-w_{01})^2 + (v-w_{02})^2 + w_{03}^2},\qquad w_{01},w_{02}\in\mathbb R^{N},\quad w_{03}\in\mathbb R.
\end{equation}
This form is expected to work if the range of the coordinates and the velocities are similar. This could need a preparation step, where the incoming $x$ and $v$ coordinates are mapped into $[-1,1]$ interval.

\subsubsection{Stability}

The second issue is about the stability of the recursion. In fact, we shall ensure that even for a large number of iteration steps the values of the position and velocity remain in the sensible range.

The core of the problem can be understood even in continuous formalism. The differential equation \eqref{eq:Newton} can be dissipative, in which case the coordinates all go to a constant value after a long time. The motion can also be divergent, in which case the velocities have larger and larger values, eventually they may also diverge in a finite time interval. Motions that remain in a finite range, i.e. they are neither dissipative, nor divergent, lie on the border line between the two extremes. But it is highly improbable that a numerically determined force lies exactly on the border line. Usually what happens that the force may have dissipative and diverging parts, and, due to the larger phase space occupied by diverging motions, practically a numerically determined force almost always leads to a divergent motion. 

The stability of these motions usually accompanied by the appearance of conserved quantities (like energy, momentum, angular momentum etc.). In numerical recursions the constancy of the conserved quantities can ensure the stability of the motion.

In more general terms we can establish $k$th order constraints (laws) in the observed data that affect $k$ level of embedding. These are:
\begin{itemize}
    \item zeroth order (holonomic) constraints are of the form
    \begin{equation}
        C^{(0)}(x_n) = C^{(0)}(x_0),\qquad \forall n.
    \end{equation}
    This means that the used coordinates are interdependent. We can avoid dealing with zeroth-order constraints if we choose independent parametrization of the observed motion. For example, a rod means a constraint $x_n^2+y_n^2 = R^2 = x_0^2 + y_0^2$ for its endpoints, this can be avoided by working with the declination angle.
    \item first-order constraints are of the form
    \begin{equation}
        C^{(1)}_{\Delta t}(x_n, v_n) = C^{(1)}_{\Delta t}(x_0, v_0)\qquad \forall n.
    \end{equation}
    These include the anholonomic constraints (in this case $C_1$ is linear in  $v$), as well as the conserved quantities.
    \item second-order constraints are of the form
    \begin{equation}
        C^{(2)}_{\Delta t}(x_n,v_n,a_n) = C^{(2)}_{\Delta t}(x_0,v_0,a_0),\qquad \forall n.
    \end{equation}
    In mechanics, second-order constraints yield complete information about the motion. The usual form is to express the $a_n$ variable, and arrive at the Newton-equation form of the equations of motion \eqref{eq:Newton}.
\end{itemize}
In a consistent system, the higher-order constraints are compatible with the lower-order ones. In particular, in mechanics, the equations of motion should respect the holonomic, anholonomic constraints as well as conserved quantities.

In the case of numerically determined force, the force usually does not support an exact conserved quantity. Then the force and the conservation are practically independent, thus we should determine the different order constraints independently, from the data. Then we require the fulfillment of all the constraints (including the equation of motion), to the precision they describe the data.

\subsection{Training}

The training set is a collection of the observed trajectories with different initial conditions. We prepare the $x_n$, $v_n$ coordinates from all the observed time series using \eqref{eq:va}, this makes the items in our dataset (our "phase space"). We also compute the acceleration proxies $a_n$ from \eqref{eq:va}, this makes the "labels" of the data. The number of all data $N_{data}$ is the number of all observed $(x_n,v_n,a_n)$ triplets.

As the Extreme Learning Machine ideas suggest, we choose the $w_0$ weights randomly and do not train them. In the choice of $w_{01}$ and $w_{02}$, we should take care that the generated features cover the observed phase space more-or-less uniformly. The $w_{03}$ variables we keep fixed for all the features.

Once we have the $w_0$ values, we also can compute the features, these are common for all observables. We denote the features by
\begin{equation}
    F_{ni} = h_i(x_n,v_n, w_0),\quad i=\{1,\dots,N_{feat}\}
\end{equation}
where $N_{feat}$ is the number of the features.

\subsubsection{Conserved quantities}

We shall choose the $w$ variables separately to obtain conserved quantities and the observed acceleration values. The conserved quantities are formed as
\begin{equation}
    C_n = \sum_{i=1}^{N_{feat}} F_{ni} w_i.
\end{equation}
This can be written in matrix form as
\begin{equation}
    \label{eq:matrixform}
    C = F\; w,
\end{equation}
where $F$ now represents a matrix. Conservation means
\begin{equation}
    C_n = C_{n+k}
\end{equation}
with arbitrary $k$. In the actual calculations, we fix the separation index. We must take care that we compare $C$ values belonging to the same trajectory, so the above condition does not apply on some $n$-s, where $n+k$ already points to the next trajectory.

To determine the $w$ weights we prepare the matrix
\begin{equation}
    dF_{ni} = F_{ni} - F_{n+k,i},
\end{equation}
and require
\begin{equation}
    |dF\;  w|^2=\mathrm{minimal,\;where\;} |w|^2=1.
\end{equation}
This is a minimum value problem with a constraint: to handle it we shall use the Lagrange multiplier method. Then we shall minimize
\begin{equation}
    L = w^T dF^T dF w -\lambda w^Tw,
\end{equation}
where $\lambda$ is to be determined. This leads to
\begin{equation}
    dF^T dF w = \lambda w
\end{equation}
eigenvalue equation. The quantity we need to minimize is then
\begin{equation}
    |dF\;  w|^2= \lambda^2=\mathrm{minimal},
\end{equation}
thus we have to choose the smallest eigenvalue. This method is analogous to the PCA method, but here we need the smallest eigenvalue, not the largest one.

We also note that in practice not only the constancy of the eigenvalues is important, but also the ability to make a distinction between motions.

\subsubsection{Force}

The force is also a linear combination of the features, like the conserved quantities. In matrix form is reads
\begin{equation}
    \label{eq:matrixform1}
    f = F\,\bar w.
\end{equation}
To determine $\bar w$ we require that the calculated force reproduces the observed accelerations the best:
\begin{equation}
    | F\,\bar w - a|^2=\mathrm{minimal}.
\end{equation}
This leads to the equation
\begin{equation}
    F^T F \;\bar w = F^T a.
\end{equation}
Although it seems that a matrix inversion could be used here, in fact usually $F^TF$ is an ill-conditioned matrix. We shall use a pseudoinverse instead; for that, we solve the eigenvalue problem for $F^TF$
\begin{equation}
    F^T F\; r^{(\alpha)} = \lambda^{(\alpha)} r^{(\alpha)},
\end{equation}
and then
\begin{equation}
    \bar w = \sum_{\alpha\in A} \frac1{\lambda^{(\alpha)}} r^{(\alpha)}(r^{(\alpha)}\cdot a),\qquad \mathrm{where}\quad\forall \alpha\in A\quad \frac{\lambda_\alpha}{\lambda_{max}} > \varepsilon.
\end{equation}
In practice, we use $\varepsilon=10^{-10}$.

\subsection{Recursion}

Once we trained the network, we can do the recursion to produce the time series. The recursion goes with the equation \eqref{eq:recursion1}. As we have already discussed, in itself it is not enough: it leads to a divergent behavior since the numerically determined force does not support exact conserved quantities. Therefore after each EoM step we also perform a projection to the conserved value surface. It is manifested in the code that we single out random directions, and we make one step towards the correct value of the conserved quantity (if needed, we decrease the step size). We repeat this stochastic minimum finding until we are closer to the desired value than some (2-3) times the standard deviation of the constancy of the given conserved quantity.

\subsection{Chaoticity}

A mechanical system with a relatively small number of degrees of freedom can lead to (usually leads to) chaotic behavior. This means that small differences between initial conditions grow exponentially.

This also means that no numerical method can yield the solution for long times. This can be checked easily, for example comparing the results of two solution methods in Python differential equation solver. With any precision, two methods yield similar results only for a restricted period of time.

This means that we can not expect that the solution of the recursion, provided by the neural network, will be close to the solution of the observed data. In this case, the only measure of the motion is the faithfulness of the force function representation as well as the stability of the conserved quantities.

\subsection{Renormalization}

The same motion can be sampled with different $\Delta t$ time separations. As it was discussed, a recursion with the discrete force supplemented by numerically determining conserved quantities is always sufficient to reconstruct the data. This means that we can determine $f_{\Delta t}$ for each $\Delta t$: this dependence is called renormalization \cite{wilson, wetterich, gies}.

In case the force is characterized by numerically determining weights, we can obtain $w_{\Delta t}$ dependence. These are known as "running" parameters in the terminology of the renormalization group.

In principle, we can fit a function to reconstruct the $w_{\Delta t}$ weights if they are smooth enough. This makes it possible to give a hint for different $\Delta t$ values. When it is known, the recursion can be accelerated significantly, since when it is possible, we can change a larger $\Delta t$.

In this work we do not use the renormalization to improve our methods, it is the task of future investigations.

\section{Studies}
\label{sec:studies}

In this section, we describe the different systems that are studied with the method described above.

\subsection{Linear oscillator}

The most simple system is the linear oscillator. In this case, we can follow the algorithm analytically.

The equation of motion, rescaled, reads
\begin{equation}
    \ddot x = -\omega^2 x,
\end{equation}
having the solution
\begin{equation}
    \label{eq:harmonic_oscillator_solution}
    x(t) = A \sin(\omega t +\phi_0).
\end{equation}

If we perform observation in $\Delta t$ time steps, then we can establish the following relation
\begin{equation}
    x(t+\Delta t) + x(t-\Delta t) = 2 \cos(\omega\Delta t) x(t).
\end{equation}
This corresponds to the recursion
\begin{equation}
    x_{n+1} = 2\cos(\omega\Delta t) x_n - x_{n-1}.
\end{equation}
expressed through $v_n$ and $a_n$ this reads
\begin{equation}
    a_n = 2\frac{\cos(\omega\Delta t)-1}{\Delta t^2} x_n.
\end{equation}
At finite step sizes, the force remains linear and independent of the velocity:
\begin{equation}
    f_{\Delta t}(x,v) = -Z(\omega\Delta t)\, \omega^2 x,
\end{equation}
where
\begin{equation}
    Z(k) = 2\frac{1-\cos k}{k^2}
\end{equation}
is the scale dependent "multiplicative mass renormalization factor". For small $k$ it is $Z(k) = 1-k^2/12+\dots$.

We can also find a conserved quantity for any scale. In the continuous case the energy
\begin{equation}
    E = \frac 12 {\dot x}^2 + \frac12 \omega^2 x^2
\end{equation}
is conserved. Using the solution \eqref{eq:harmonic_oscillator_solution} we find that
\begin{equation}
    {\dot x}(t) = A\omega \cos(\omega t +\phi_0).
\end{equation}
On the other hand with $k=\omega \Delta t$ from \eqref{eq:harmonic_oscillator_solution} we find
\begin{equation}
    x(t-\Delta t) = x(t) \cos k - A \cos(\omega t +\phi_0)\sin k,
\end{equation}
therefore
\begin{equation}
    {\dot x}(t) = \omega\frac{x(t) \cos k -x(t-\Delta t)}{\sin k}.
\end{equation}
Using $t=n\Delta t$ and \eqref{eq:va} we find
\begin{equation}
    {\dot x}(n\Delta t) = \omega\frac{x_n \cos k -x_{n-1}}{\sin k} = \frac{v_n k - \omega x_n (1-\cos k)}{\sin k}.
\end{equation}
Finally, we obtain
\begin{equation}
    C^{(1)}_{\Delta t}(x,v) = \frac12 \left(\frac{k v- \omega x(1- \cos k)}{\sin k}\right)^2 +\frac12\omega^2 x^2.
\end{equation}
In the $k\to0$ limit, the first term indeed becomes simply $\frac12 {\dot x}^2$.

\subsection{Gravity pendulum}

A slightly more complicated system is the pendulum in the gravitational field. In the simplest version it is governed by the EoM (after appropriate rescaling of the time variable)
\begin{equation}
    \label{eq:gravity_pendulum_eom}
    \ddot x = -\sin x.
\end{equation}
The physical meaning of $x$ is the deflection angle.

This differential equation has a conserved quantity (energy)
\begin{equation}
    E = \frac12 \dot x^2 +1 - \cos x.
\end{equation}
This quantity is bounded from below ($E>0$). For $E<2$ ($|\dot x|<2$) the motion is periodic, the maximal angle is $1-E = \cos x$.

\begin{figure}
    \centering
    \includegraphics[width=12cm]{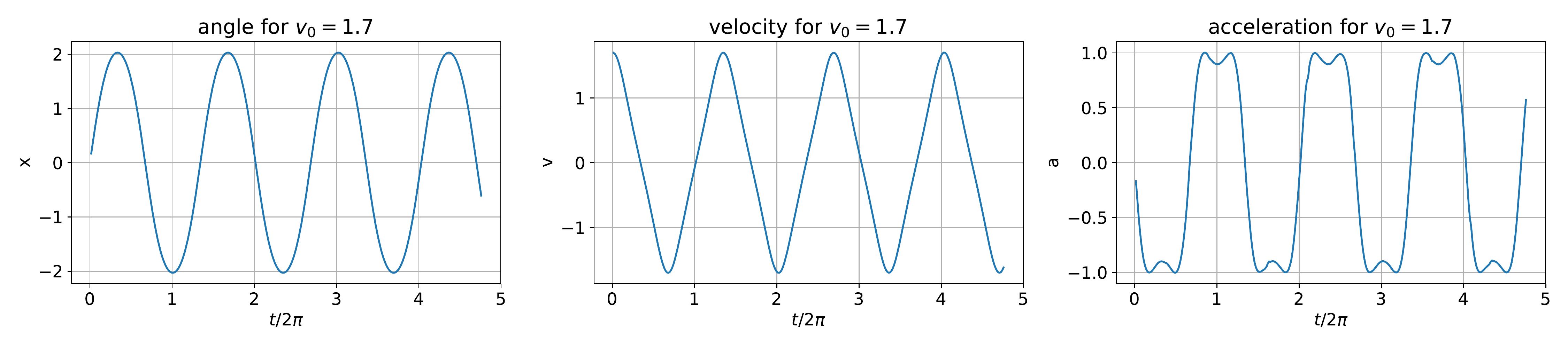}
    \caption{Solution of the gravity pendulum for near non-periodic initial conditions. The period in small amplitude approximation is $2\pi$.}
    \label{fig:gravpend_solution}        
\end{figure}

We can solve \eqref{eq:gravity_pendulum_eom} numerically. An example for close-to non-periodic motion can be seen in Figure.~\ref{fig:gravpend_solution}.
The period in the small amplitude case is $2\pi$, we normalized the time variable with this value.

We prepare our network following the general descriptions in subsection \ref{sec:network}. We used $N_{feat}=100$ features in the hidden layer, the $w_{01}, w_{02}$ position variables were uniformly distributed in the range where the training set data was present, the scale variable was $w_{03}=2$. For the training we used simulated data with initial conditions $x(0)=0$ and $v(0)=0.5, 1.7, 2.2, 2.6$ and $3$. The first two are periodic, the last three are non-periodic motions. For the data set we used a discretization time $\Delta t = 0.1$. We note that this value is rather large, where the typical scale of change is of order one.

\begin{figure}
    \centering
    \includegraphics[width=7cm]{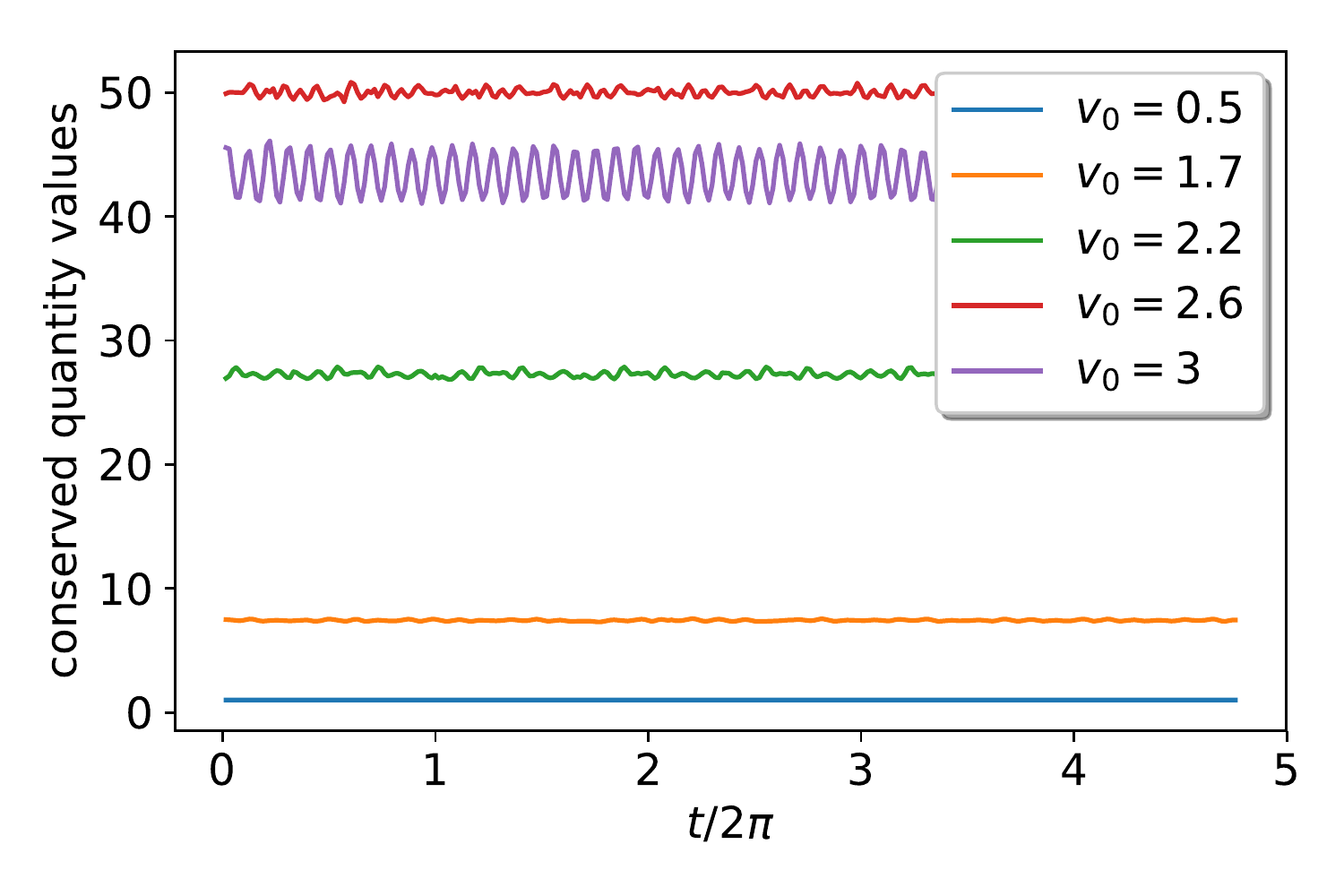}
    \caption{Average of the conserved quantities in the gravity pendulum case, normalized. The precision of conservation is about $2.7\times 10^{-3}$.}
    \label{fig:gravpend_conserved}        
\end{figure}

We determined two conserved quantities, their average can be seen in Figure~\ref{fig:gravpend_conserved}.
We see that their values are fluctuating, but the average remains the same over time.

\begin{figure}
    \centering
    \includegraphics[width=7cm]{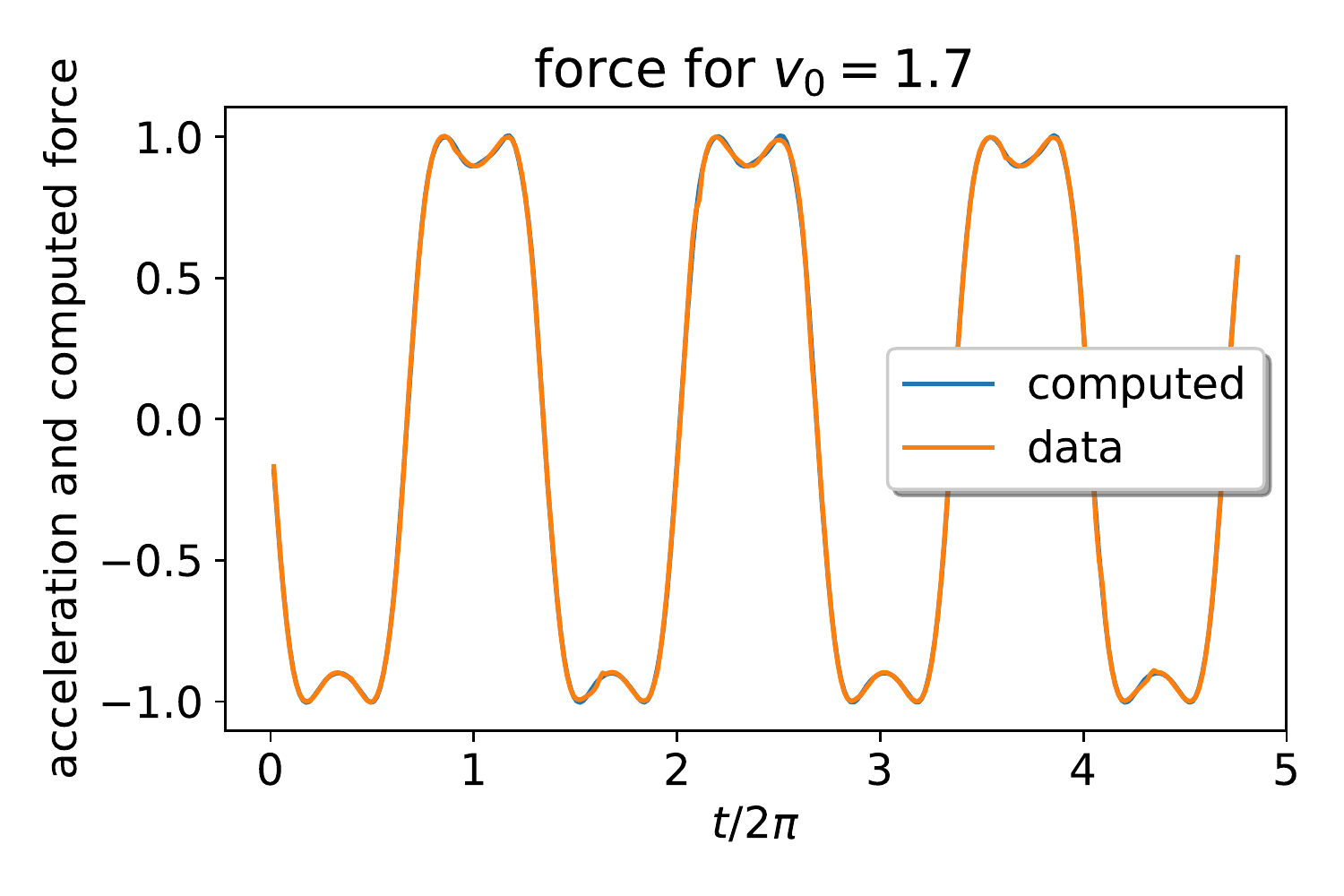}
    \caption{The observed acceleration and the computed force for the gravity pendulum case with $v_0=1.7$. The precision of the reproduction of the force function is $96.6$\%.}
    \label{fig:gravpend_force}        
\end{figure}

We also trained the force. The value of the force along the trajectory of the $v_0=1.7$ motion and the computed value is shown in Figure~\ref{fig:gravpend_force}.
As we see, the computed force and the measured acceleration is hardly different. Even where they are not the same, its reason is that the numerical solver produces sometimes non-continuous acceleration. The precision of the reproduction of the force was $96.6\%$.

\begin{figure}
    \centering
    \includegraphics[width=7cm]{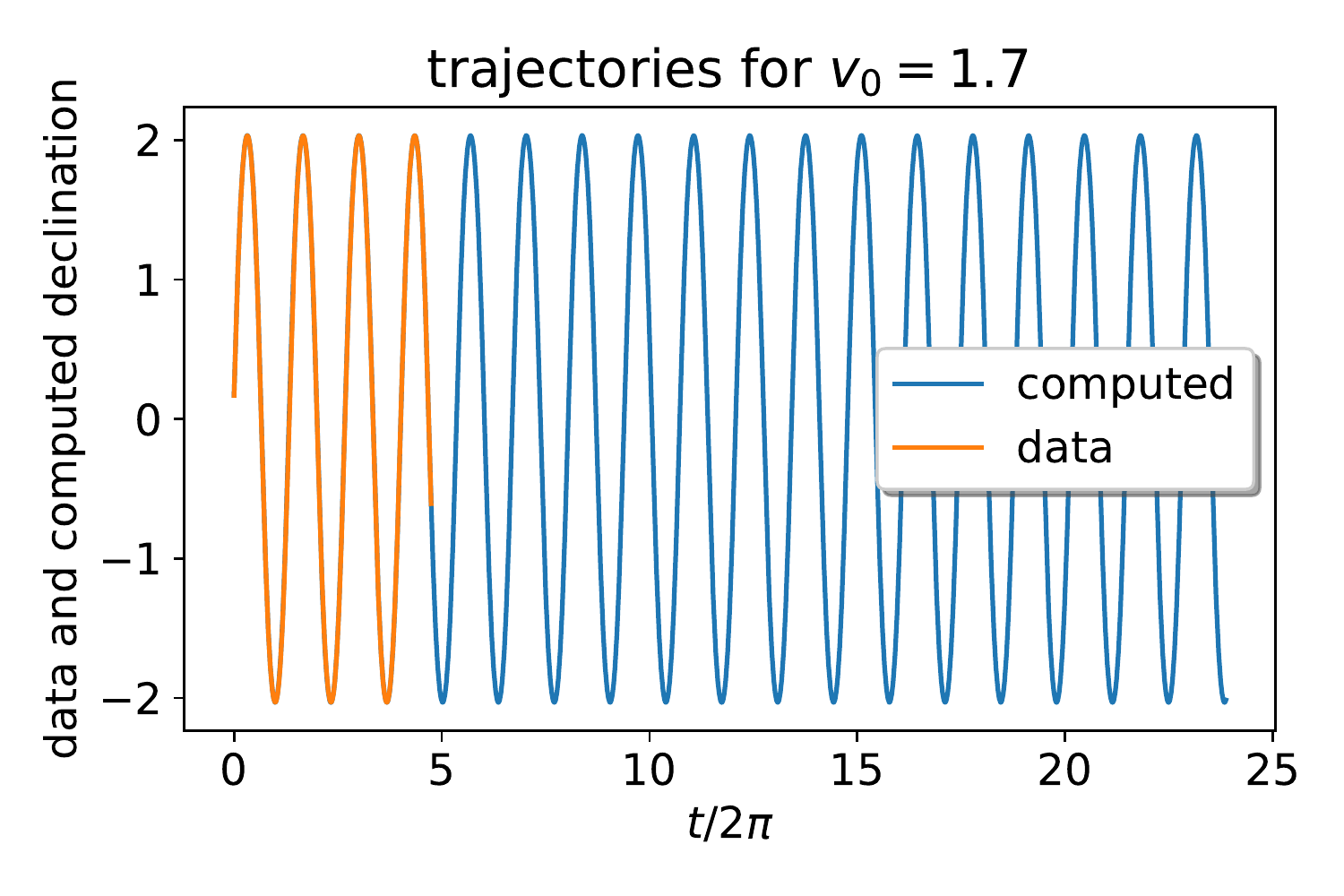}
    \caption{The reconstructed trajectory together with the original one.}
    \label{fig:gravpend_motion}        
\end{figure}

After the training, we may run a recursion. We continued the motion for 10 times the training period, and the result can be seen in Figure~\ref{fig:gravpend_motion}.
As we can see, the reconstructed and the original trajectories are hardly distinguishable (in the figure they are within line width distance). Quantitatively, the reconstruction error is $0.83$\%.

\subsection{Double pendulum}

Using the same ideas we can work out the motion of the more complicated double pendulum. As a differential equation, its equations of motion read
\begin{eqnarray}
    \ddot  x_1 = && -\frac1{l_1 (M - m_2 \cos^2 \Delta\phi^2)}\biggl(g M \sin x_1
    - g m_2 \cos\Delta\phi \sin x_2 +\nonumber\\  &&
    + \frac 12 l_1 m_2  \dot x_1^2 \sin 2\Delta\phi + 
    l_2 m_2 \dot x_2^2\sin \Delta\phi\biggr),\nonumber\\
    \ddot  x_2 = && \frac{\sin \Delta\phi }{l_2 (M - m_2 \cos^2 \Delta\phi^2)} \biggl(g M \cos  x_1 + l_1 M \dot x_1^2 
        + l_2 m_2 \dot x_2^2 \cos\Delta\phi\biggr).
\end{eqnarray}
where $\Delta\phi= x_1- x_2$ and $M=m_1+m_2$.

\begin{figure}
    \centering
    \includegraphics[width=12cm]{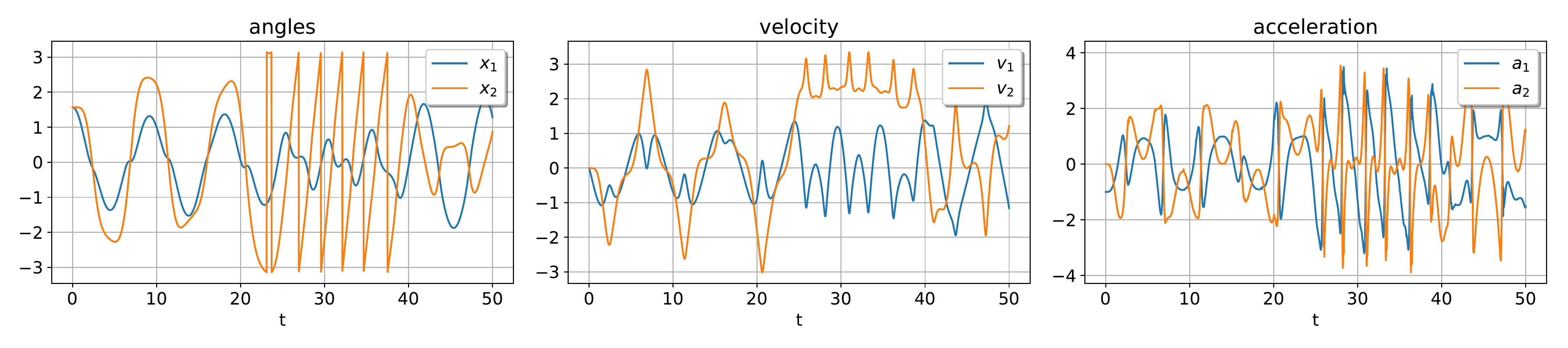}
    \includegraphics[width=12cm]{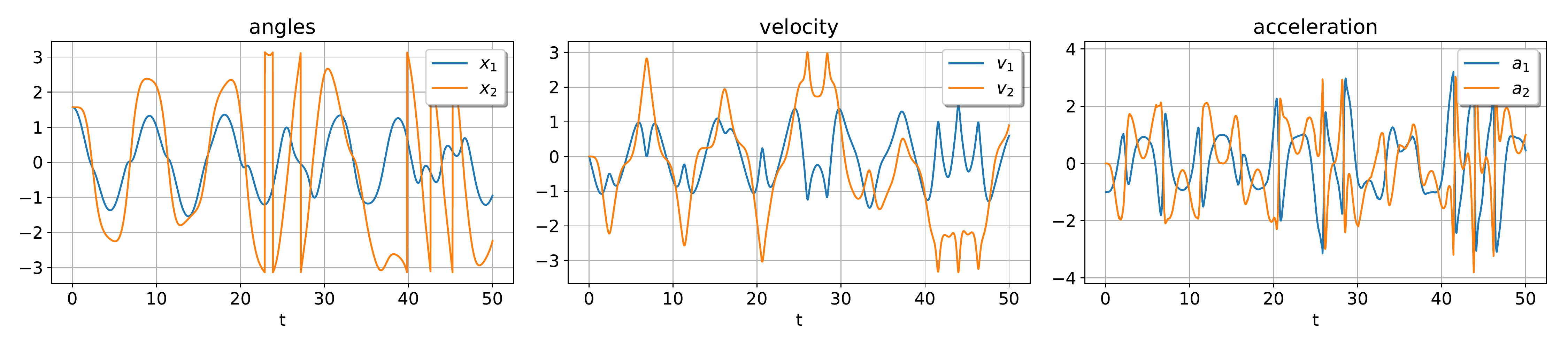}
    \caption{Trajectories coming from the double pendulum system for initial conditions $( x_1, x_2,\dot x_1,\dot x_2) = (\pi/2,\pi/2,0,0)$ for two different solver: the upper is for 'DOP853', the lower is for 'RK45' in the \texttt{solve\_ivp} function of Scipy (Python).}
    \label{fig:dbpsolution}        
\end{figure}

We first observe the motion of the double pendulum. Technically we solve the above differential equations with parameters $\ell_1=\ell_2=1,\, M=3,\,m_2=1,\,g=1$, and for three initial conditions:
\[( x_1, x_2,\dot x_1,\dot x_2) = ((\frac\pi 2,\frac\pi 2,0,0), (\frac{3\pi} 4,\frac\pi 2,0,0),(\frac\pi 4,\frac\pi 4,0,1)).\]
As an example, we show the angles, the velocities, and the acceleration in Figure~\ref{fig:dbpsolution} for the first case.
As this Figure demonstrates, the solution of the double pendulum system numerically can not be determined exactly. The reason is that this system is chaotic, and any small difference in the state of the system leads to exponentially deviating solutions. In Figure~\ref{fig:dbpsolution} we see the example of the solution using two differential equation solver methods of Python's Scipy package: the 'DOP853' and the 'RK45' methods. We see that the solution starts in a similar way, but soon they become different. This is, actually, not a bug, but a feature of chaotic systems. But this also means that we can not expect that our numerical method provides trajectories that stay close to an exact one.

After we have the observed data, we can build up and train our network. We used $N_{feat}=1000$ features in the hidden layer, the $w_{01}, w_{02}$ position variables were uniformly distributed in the range where the training set data was present, the scale variable was $w_{03}=3$. For the training, we used $\Delta t = 0.02$ time discretization. We used two conserved quantities here.

\begin{figure}
    \centering
    \includegraphics[width=12cm]{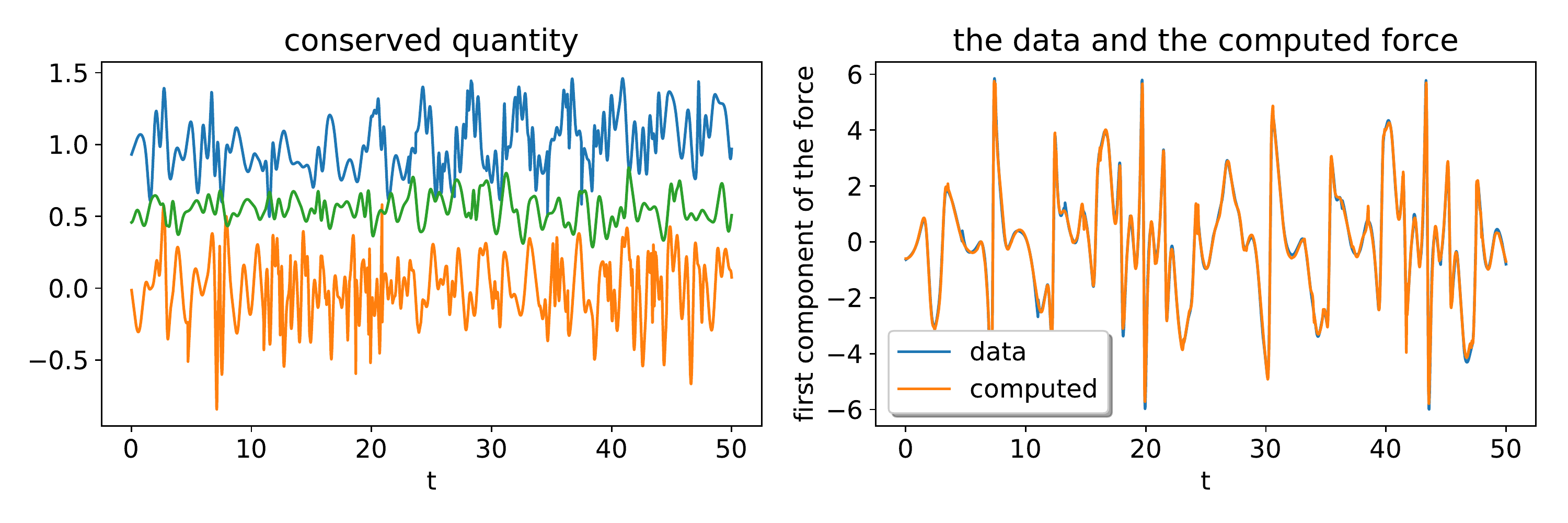}
    \caption{Left panel: time dependence of the average of the conserved quantities for the three observed trajectories. Right panel: the first component of the observed acceleration and the reconstructed force for the first motion.}
    \label{fig:dbptrained}        
\end{figure}

In Figure~\ref{fig:dbptrained} we show the time dependence of the conserved quantities and the time dependence of the acceleration and the reconstructed force.
We can see that although the conserved quantities fluctuate in time, their average remains considerably constant; this persists for later times, too. The accuracy of the force determination is 93\%; this is partly due to the differential equation solver which produces sometimes glitches in the acceleration.

\begin{figure}
    \centering
    \includegraphics[width=7cm]{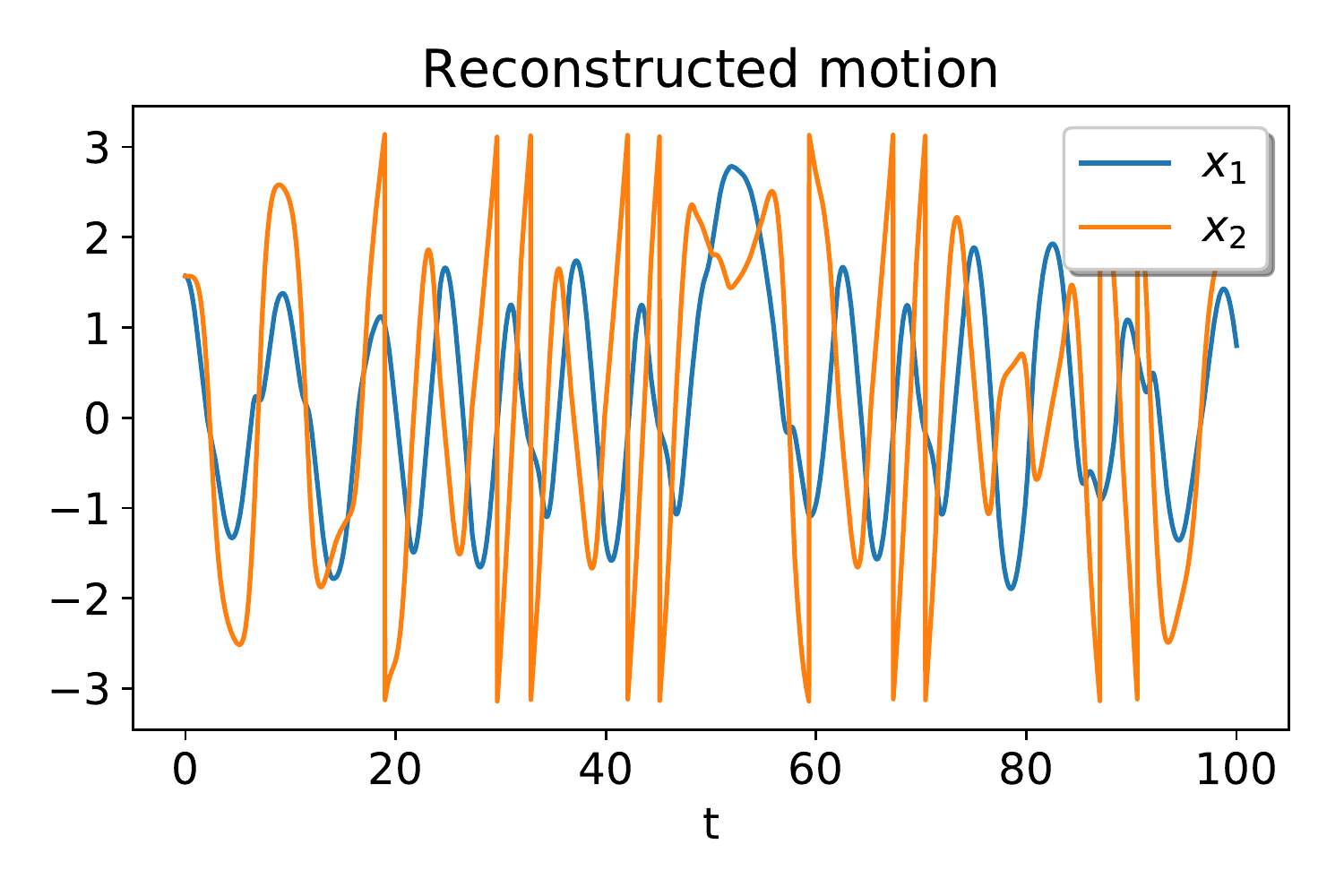}
    \caption{The reconstructed motion of the double pendulum.}
    \label{fig:reconstructed}        
\end{figure}

With the trained network we can reconstruct the motion, the result can be seen in Figure~\ref{fig:reconstructed}.
As we see, the early time behavior is the same as was in the case of both differential equation solving methods. But the deviation starts earlier: this is because the reconstructed force is less accurate than the numerical errors made by the numerical differential equation solvers. But what is important, the motion remains stable, there are no divergent parts. This property is the consequence of the requirement that conserved quantities remain conserved up to their standard deviation.

\section{Conclusion}
\label{sec:conclusion}

In this paper, we proposed a method that is able to reconstruct and continue observed mechanical motions. The input of this method is the trajectories discretized in time. The core is the module that is capable to determine the different level laws in the motion: the holonomic constraints, the anholonomic constraints, the conserved quantities, and the equation of motion. For a stable reconstruction of the motion, one has to use all of this information, since the EoM alone leads to unstable, diverging solutions.

For the representation of the laws a shallow neural network was used, the training followed the Extreme Learning Machine ideas. Here only the last layer is trained, all the former layers are thought to provide the features of the problem, which is combined linearly by the last layer. We used a relatively small number of parameters, where the hidden layer had $N_{feat}=100-1000$ elements. Therefore the number of the weights was at most some thousand. This leads to a fast learning and fast reconstruction of the motion. With this network, the observed force could be reconstructed with better than 90\% precision. 

As applications, we worked out analytically the mathematical pendulum case. The numerical method was applied to the pendulum in the gravitational field and the double pendulum case. Whenever the motion is not chaotic, our method could reconstruct the observed motion with high precision and could continue the motion in a stable way. In the chaotic case, the exact motion can not be determined numerically, since all numerical methods contain approximations, and in chaotic systems, small deviations grow exponentially in time. This also means that our numerical method could not exactly reproduce the observed data, since the representation of the observed force was not completely correct. Nevertheless, the generated motion remained stable in time, due to the requirement of the constancy of the conserved quantities.

\section*{Acknowledgment}

The authors acknowledge useful discussions with A. Telcs, T. Biro, Z. Somogyvari. The research was supported by the Ministry of Innovation and Technology NRDI Office within the
framework of the MI-LAB Artificial Intelligence National Laboratory Program. A.J. had a support from the Hungarian Research
Fund NKFIH (OTKA) under contract No. K123815.

\end{document}